# When do Numbers Really Matter?

**Hei Chan**                                                    HEI@CS.UCLA.EDU
**Adnan Darwiche**                                      DARWICHE@CS.UCLA.EDU
*Computer Science Department*
*University of California, Los Angeles*
*Los Angeles, CA 90095, USA*

## Abstract

Common wisdom has it that small distinctions in the probabilities (parameters) quantifying a belief network do not matter much for the results of probabilistic queries. Yet, one can develop realistic scenarios under which small variations in network parameters can lead to significant changes in computed queries. A pending theoretical question is then to analytically characterize parameter changes that do or do not matter. In this paper, we study the sensitivity of probabilistic queries to changes in network parameters and prove some tight bounds on the impact that such parameters can have on queries. Our analytic results pinpoint some interesting situations under which parameter changes do or do not matter. These results are important for knowledge engineers as they help them identify influential network parameters. They also help explain some of the previous experimental results and observations with regards to network robustness against parameter changes.

## 1. Introduction

A belief network is a compact representation of a probability distribution (Pearl, 1988; Jensen, 2001). It consists of two parts, one qualitative and the other quantitative. The qualitative part of a belief network (called its structure) is a directed acyclic graph in which nodes represent domain variables and edges represent direct influences between these variables. The quantitative part of a belief network is a set of conditional probability tables (CPTs) that quantify our beliefs in such influences. Figure 1 depicts the structure of a belief network and Figure 2 depicts its CPTs.[1]

Automated reasoning systems based on belief networks have become quite popular recently as they have enjoyed much success in a number of real-world applications. Central to the development of such systems is the construction of a belief network (hence, a probability distribution) that faithfully represents the domain of interest. Although the automatic synthesis of belief networks—based on design information in certain applications and based on learning techniques in others—has been drawing a lot of attention recently, mainstream methods for constructing such networks continue to be based on traditional knowledge engineering (KE) sessions involving domain experts. One of the central issues that arise in such KE sessions is the assessment of impact that changes in network parameters may have on probabilistic queries of interest.

Consider for example the following common method for constructing belief networks in medical diagnosis applications (Coupé, Peek, Ottenkamp, & Habbema, 1999). First, the network structure is developed. Next, parameters are estimated by non-experts using a combination of statistical data and qualitative influences available from textbook materials. Finally, medical experts are brought in to evaluate the network and fine-tune its parameters. One method of evaluation is to pose diagnostic scenarios to the network, and compare the results of such queries to those expected by the experts. For example, given some set of symptoms $\mathbf{e}$, and two potential diagnoses $y$ and $z$, the network may give us the conclusion that $Pr(y \mid \mathbf{e})/Pr(z \mid \mathbf{e}) = 2$, while a domain expert may believe that the ratio should be no less than 4. Assuming that the network structure is correct, a central question is then: which network parameters should be changed to give us the correct ratio, and by how much?

---

1. This specific network and its CPTs are distributed with the evaluation version of the commercial HUGIN system at `http://www.hugin.com/`.

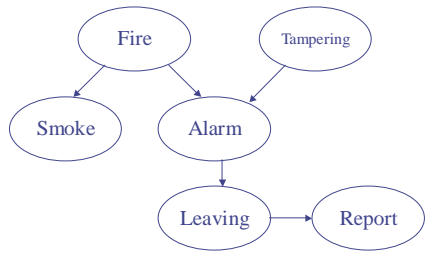

Figure 1: A belief network structure.

| | Fire | $\theta_{x\mid\mathbf{u}}$ |
|---|---|---|
| | true | .01 |
| | false | .99 |

| | Tampering | $\theta_{x\mid\mathbf{u}}$ |
|---|---|---|
| | true | .02 |
| | false | .98 |

| Fire | Smoke | $\theta_{x\mid\mathbf{u}}$ |
|---|---|---|
| true | true | .9 |
| true | false | .1 |
| false | true | .01 |
| false | false | .99 |

| Fire | Tampering | Alarm | $\theta_{x\mid\mathbf{u}}$ |
|---|---|---|---|
| true | true | true | .5 |
| true | true | false | .5 |
| true | false | true | .99 |
| true | false | false | .01 |
| false | true | true | .85 |
| false | true | false | .15 |
| false | false | true | .0001 |
| false | false | false | .9999 |

| Alarm | Leaving | $\theta_{x\mid\mathbf{u}}$ |
|---|---|---|
| true | true | .88 |
| true | false | .12 |
| false | true | .001 |
| false | false | .999 |

| Leaving | Report | $\theta_{x\mid\mathbf{u}}$ |
|---|---|---|
| true | true | .75 |
| true | false | .25 |
| false | true | .01 |
| false | false | .99 |

Figure 2: The CPTs of the belief network shown in Figure 1.

To automate the task of identifying such parameter changes, we have recently developed a belief network tool, called SAMIAM (Sensitivity Analysis, Modelling, Inference And More)[2]. One of its feature is sensitivity analysis, which allows domain experts to fine-tune network parameters in order to enforce constraints on the results of certain queries. Users can specify the constraint that they want to enforce, and SamIam will automatically decide whether a given parameter is relevant to this constraint, and if it is, will compute the minimum amount of change to that parameter which is needed to enforce the constraint. The technical details of our approach to sensitivity analysis are the subject of Section 2.

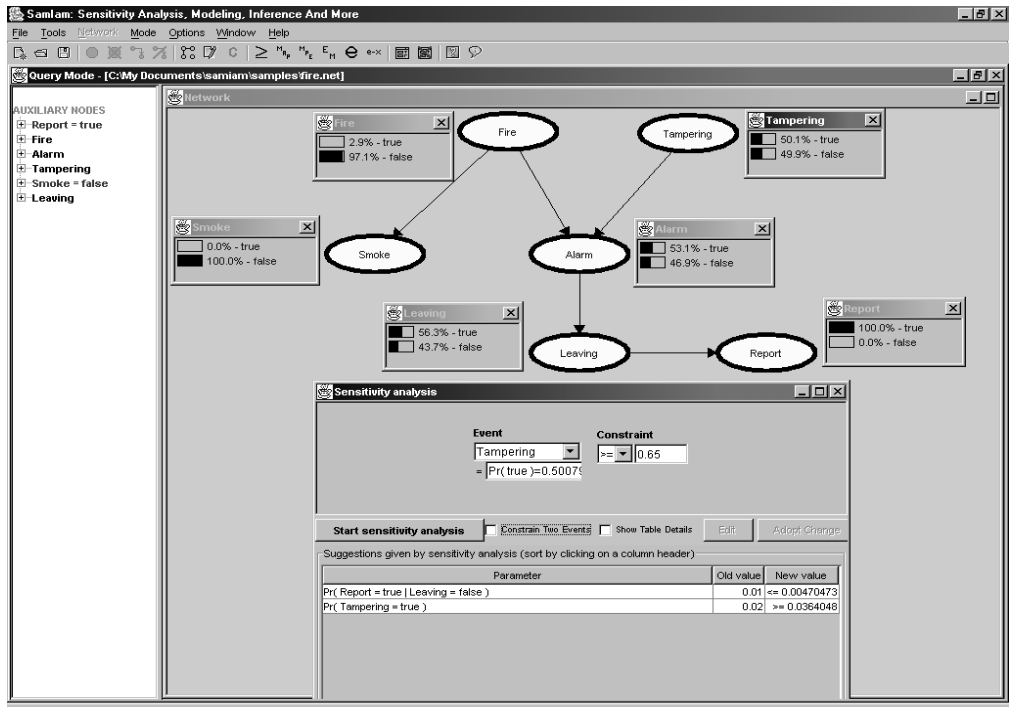

Figure 3: A screen shot of SamIam performing sensitivity analysis on the belief network shown in Figure 1.

As we experimented with SamIam, we ran into scenarios that we found to be surprising at first glance. Specifically, there were many occasions in which queries would be quite sensitive to small variations in certain network parameters. Consider the scenario in Figure 3 for one example, which corresponds to the network detailed in Figures 1 and 2. Here, we have evidence $\mathbf{e} = report, \overline{smoke}$: people are reported to be evacuating a building, but there is no evidence for any smoke. This evidence should make *tampering* more likely than *fire*, and the given belief network does indeed reflect this with $Pr(tampering \mid \mathbf{e}) = .50$ and $Pr(fire \mid \mathbf{e}) = .03$. We wanted, however, the probability of *tampering* to be no less than .65. Hence, we asked SamIam to identify parameter changes that can enforce this constraint, and it made two recommendations:

1. either decrease the probability of a false report, $Pr(report \mid \overline{leaving})$, from its current value of .01 to $\leq .0047$,

2. or increase the prior probability of *tampering* from its current value of .02 to $\geq$ .036.

Therefore, the distinctions between .02 and .036, and the one between .01 and .0047, do really matter in this case as each induces an absolute change of .15 on the probabilistic query of interest. Note also that implicit in SAMIAM's recommendations is that the parameters of variables *Fire*, *Smoke*, *Leaving*, and *Alarm* are irrelevant to enforcing this constraint, i.e. no matter how much we change any of these parameters, we would not be able to enforce our desired constraint.

This example shows that the *absolute change* in a query can be much larger than the *absolute change* in the corresponding parameters. Later, we will show an example where an infinitesimal change to a network parameter leads to a change of .5 to a corresponding query. We also show examples in which the *relative change* in the probability of a query is larger than the corresponding *relative change* in a network parameter. One wonders then whether there is a different method for measuring probabilistic change (other than absolute or relative), which allows one to non-trivially bound the change in a probabilistic query in terms of the corresponding change in a network parameter.

To answer this and related questions, we conduct in Section 3 an analytic study of the partial derivative of a probabilistic query $Pr(y \mid \mathbf{e})$ with respect to some network parameter $\theta_{x|\mathbf{u}}$. Our study leads us to three main results:

1. a bound on the derivative in terms of $Pr(y \mid \mathbf{e})$ and $Pr(x \mid \mathbf{u})$ only, which is independent of any other aspect of the given belief network;

2. a bound on the sensitivity of queries to infinitesimal changes in network parameters;

3. a bound on the sensitivity of queries to arbitrary changes in network parameters.

The last bound in particular shows that the amount of change in a probabilistic query can be bounded in terms of the amount of change in a network parameter, as long as change is understood to be the *relative change in odds*. This result has a number of practical implications. First, it can relieve experts from having to be too precise when specifying certain parameters subjectively. Next, it can be important for approximate inference algorithms that pre-process network parameters to eliminate small distinctions in such parameters, in order to increase the efficiency of inference (Poole, 1998). Finally, it can be used to show that automated reasoning systems based on belief networks are robust and, hence, suitable for real-world applications (Pradhan, Henrion, Provan, Del Favero, & Huang, 1996).

Section 4 is indeed dedicated to exploring the implications of the above bounds, where we provide an analytic explanation of why certain parameter changes don't matter. We finally close in Section 5 with some concluding remarks. Proofs of all theorems are given in Appendix A.

## 2. The Tuning of Network Parameters

We report in this section on a tool that we have been developing, called SAMIAM, for fine-tuning network parameters (Laskey, 1995; Castillo, Gutiérrez, & Hadi, 1997; Jensen, 1999; Kjærulff & van der Gaag, 2000; Darwiche, 2000). Given a belief network, some evidence $\mathbf{e}$, which is an instantiation of variables $\mathbf{E}$ in the belief network, and two events $y$ and $z$ of variables $Y$ and $Z$ respectively, where $Y, Z \notin \mathbf{E}$, our tool can efficiently identify parameter changes needed to enforce the following types of constraints:

DIFFERENCE: $Pr(y \mid \mathbf{e}) - Pr(z \mid \mathbf{e}) \geq \epsilon$;

RATIO: $Pr(y \mid \mathbf{e})/Pr(z \mid \mathbf{e}) \geq \epsilon$.

These two constraints often arise when we debug belief networks. For example, we can make event $y$ more likely than event $z$, given evidence $\mathbf{e}$, by specifying the constraint, $Pr(y \mid \mathbf{e}) - Pr(z \mid \mathbf{e}) \geq 0$, or we can make event $y$ at least twice as likely as event $z$, given evidence $\mathbf{e}$, by specifying the constraint, $Pr(y \mid \mathbf{e})/Pr(z \mid \mathbf{e}) \geq 2$. We will discuss next how one would enforce the two constraints, but we need to settle some notational conventions and technical preliminaries first.

Variables are denoted by upper-case letters ($A$) and their values by lower-case letters ($a$). Sets of variables are denoted by bold-face upper-case letters ($\mathbf{A}$) and their instantiations are denoted by bold-face lower-case letters ($\mathbf{a}$). For a variable $A$ with values *true* and *false*, we use $a$ to denote $A = true$ and $\overline{a}$ to denote $A = false$. The CPT for variable $X$ with parents $\mathbf{U}$ defines a set of conditional probabilities of the form $Pr(x \mid \mathbf{u})$, where $x$ is a value of variable $X$, $\mathbf{u}$ is an instantiation of parents $\mathbf{U}$, and $Pr(x \mid \mathbf{u})$ is a probability known as a *network parameter* and denoted by $\theta_{x|\mathbf{u}}$. We finally recall a basic fact about belief networks. The probability of some instantiation $\mathbf{x}$ of all network variables $\mathbf{X}$ equals the product of all network parameters that are consistent with that instantiation. For example, the probability of instantiation *fire*, $\overline{tampering}$, *smoke*, *alarm*, $\overline{leaving}$, *report* in Figure 1 equals $.01 \times .98 \times .9 \times .99 \times .12 \times .01$, which is the product of network parameters (from Figure 2) that are consistent with this instantiation.

## 2.1 Binary Variables

We first consider the parameters of a binary variable $X$, with two values $x$ and $\overline{x}$ and, hence, two parameters $\theta_{x|\mathbf{u}}$ and $\theta_{\overline{x}|\mathbf{u}}$ for each parent instantiation $\mathbf{u}$. We assume that for each variable $X$ and parent instantiation $\mathbf{u}$ we have a meta parameter $\tau_{x|\mathbf{u}}$, such that $\theta_{x|\mathbf{u}} = \tau_{x|\mathbf{u}}$ and $\theta_{\overline{x}|\mathbf{u}} = 1 - \tau_{x|\mathbf{u}}$. Therefore, our goal is then to determine the amount of change to the meta parameter $\tau_{x|\mathbf{u}}$ which would lead to a simultaneous change in both $\theta_{x|\mathbf{u}}$ and $\theta_{\overline{x}|\mathbf{u}}$. We use the meta parameter $\tau_{x|\mathbf{u}}$ because it is not meaningful to change only $\theta_{x|\mathbf{u}}$ or $\theta_{\overline{x}|\mathbf{u}}$ without changing the other since $\theta_{x|\mathbf{u}} + \theta_{\overline{x}|\mathbf{u}} = 1$.

First we observe that the probability of an instantiation $\mathbf{e}$, $Pr(\mathbf{e})$, is a linear function in any network parameter $\theta_{x|\mathbf{u}}$ in a belief network (Russell, Binder, Koller, & Kanazawa, 1995; Castillo et al., 1997). In fact, the probability is linear in any meta parameter $\tau_{x|\mathbf{u}}$:

**Theorem 2.1** *The derivative of $Pr(\mathbf{e})$ with respect to the meta parameter $\tau_{x|\mathbf{u}}$ is given by:*

$$\frac{\partial Pr(\mathbf{e})}{\partial \tau_{x|\mathbf{u}}} = \frac{Pr(\mathbf{e}, x, \mathbf{u})}{\theta_{x|\mathbf{u}}} - \frac{Pr(\mathbf{e}, \overline{x}, \mathbf{u})}{\theta_{\overline{x}|\mathbf{u}}}, \tag{1}$$

*when $\theta_{x|\mathbf{u}} \neq 0$ and $\theta_{\overline{x}|\mathbf{u}} \neq 0$.[3] We will designate the derivative as constant $\alpha_{\mathbf{e}}$.*

In Theorem 2.1, $\alpha_{\mathbf{e}} = Pr(\mathbf{e}, x, \mathbf{u})/\theta_{x|\mathbf{u}} - Pr(\mathbf{e}, \overline{x}, \mathbf{u})/\theta_{\overline{x}|\mathbf{u}}$ is a constant in terms of both $\theta_{x|\mathbf{u}}$ and $\theta_{\overline{x}|\mathbf{u}}$ (and consequently, $\tau_{x|\mathbf{u}}$) since $Pr(\mathbf{e}, x, \mathbf{u}) = K_x \theta_{x|\mathbf{u}}$ and $Pr(\mathbf{e}, \overline{x}, \mathbf{u}) = K_{\overline{x}} \theta_{\overline{x}|\mathbf{u}}$,

---

3. If either of the previous parameters is zero, we can use the differential approach by Darwiche (2000) to compute the derivative directly.

where $K_x = Pr(\mathbf{u})Pr(\mathbf{e} \mid x, \mathbf{u})$ and $K_{\overline{x}} = Pr(\mathbf{u})Pr(\mathbf{e} \mid \overline{x}, \mathbf{u})$ are constants in terms of both $\theta_{x|\mathbf{u}}$ and $\theta_{\overline{x}|\mathbf{u}}$. By substituting $y, \mathbf{e}$ and $z, \mathbf{e}$ for $\mathbf{e}$ in Theorem 2.1, we get:

$$\alpha_{y,\mathbf{e}} = \frac{\partial Pr(y, \mathbf{e})}{\partial \tau_{x|\mathbf{u}}} = \frac{Pr(y, \mathbf{e}, x, \mathbf{u})}{\theta_{x|\mathbf{u}}} - \frac{Pr(y, \mathbf{e}, \overline{x}, \mathbf{u})}{\theta_{\overline{x}|\mathbf{u}}}; \tag{2}$$

$$\alpha_{z,\mathbf{e}} = \frac{\partial Pr(z, \mathbf{e})}{\partial \tau_{x|\mathbf{u}}} = \frac{Pr(z, \mathbf{e}, x, \mathbf{u})}{\theta_{x|\mathbf{u}}} - \frac{Pr(z, \mathbf{e}, \overline{x}, \mathbf{u})}{\theta_{\overline{x}|\mathbf{u}}}. \tag{3}$$

Now, if we want to enforce the DIFFERENCE constraint, $Pr(y \mid \mathbf{e}) - Pr(z \mid \mathbf{e}) \geq \epsilon$, it suffices to ensure that $Pr(y, \mathbf{e}) - Pr(z, \mathbf{e}) \geq \epsilon Pr(\mathbf{e})$. Suppose that the previous constraint does not hold, and we wish to establish it by applying a change of $\delta$ to the meta parameter $\tau_{x|\mathbf{u}}$. Such a change leads to a change of $\alpha_{\mathbf{e}}\delta$ in $Pr(\mathbf{e})$. It also changes $Pr(y, \mathbf{e})$ and $Pr(z, \mathbf{e})$ by $\alpha_{y,\mathbf{e}}\delta$ and $\alpha_{z,\mathbf{e}}\delta$, respectively. Hence, to enforce the DIFFERENCE constraint, we need to solve for $\delta$ in the following inequality:

$$[Pr(y, \mathbf{e}) + \alpha_{y,\mathbf{e}}\delta] - [Pr(z, \mathbf{e}) + \alpha_{z,\mathbf{e}}\delta] \geq \epsilon[Pr(\mathbf{e}) + \alpha_{\mathbf{e}}\delta].$$

Rearranging the terms, we get the following result.

**Corollary 2.1** *To satisfy the* DIFFERENCE *constraint, we need to change the meta parameter $\tau_{x|\mathbf{u}}$ by $\delta$, such that:*

$$Pr(y, \mathbf{e}) - Pr(z, \mathbf{e}) - \epsilon Pr(\mathbf{e}) \geq \delta[-\alpha_{y,\mathbf{e}} + \alpha_{z,\mathbf{e}} + \epsilon\alpha_{\mathbf{e}}],$$

*where the $\alpha$ constants are defined by Equations 1, 2 and 3.*

We can similarly solve for parameter changes $\delta$ that enforce the RATIO constraint, $Pr(y \mid \mathbf{e})/Pr(z \mid \mathbf{e}) \geq \epsilon$, in the following inequality:

$$[Pr(y, \mathbf{e}) + \alpha_{y,\mathbf{e}}\delta]/[Pr(z, \mathbf{e}) + \alpha_{z,\mathbf{e}}\delta] \geq \epsilon.$$

Rearranging the terms, we get the following result.

**Corollary 2.2** *To satisfy the* RATIO *constraint, we need to change the meta parameter $\tau_{x|\mathbf{u}}$ by $\delta$, such that:*

$$Pr(y, \mathbf{e}) - \epsilon Pr(z, \mathbf{e}) \geq \delta[-\alpha_{y,\mathbf{e}} + \epsilon\alpha_{z,\mathbf{e}}],$$

*where the $\alpha$ constants are defined by Equations 2 and 3.*

For both the DIFFERENCE and RATIO constraints, the solution of $\delta$, if any, is always in one of two forms:

- $\delta \leq q$, for some computed $q < 0$, in which case the new value of meta parameter $\tau_{x|\mathbf{u}}$ must be in the interval $[0, p + q]$.

- $\delta \geq q$, for some computed $q > 0$, in which case the new value of meta parameter $\tau_{x|\mathbf{u}}$ must be in the interval $[p + q, 1]$.

Note that $p$ is the current value of meta parameter $\tau_{x|\mathbf{u}}$ (before the change). For many parameters, these intervals are empty and, therefore, there is no way we can change these meta parameters to enforce the constraint.

The question now is how to solve these inequalities, efficiently, and for all meta parameters. Note that there may be more than one possible parameter change that would enforce the given constraint, so we need to identify all such changes. With either Corollary 2.1 or 2.2, we can easily solve for the amount of change needed, $\delta$, once we know the following probabilities: $Pr(\mathbf{e})$, $Pr(y, \mathbf{e})$, $Pr(z, \mathbf{e})$, $Pr(\mathbf{e}, x, \mathbf{u})$, $Pr(\mathbf{e}, \overline{x}, \mathbf{u})$, $Pr(y, \mathbf{e}, x, \mathbf{u})$, $Pr(y, \mathbf{e}, \overline{x}, \mathbf{u})$, $Pr(z, \mathbf{e}, x, \mathbf{u})$, and $Pr(z, \mathbf{e}, \overline{x}, \mathbf{u})$. This leads to the following complexity of our technique.

**Corollary 2.3** *If we have an algorithm that can compute $Pr(\mathbf{i}, x, \mathbf{u})$, for a given instantiation $\mathbf{i}$, and all family instantiations $x, \mathbf{u}$ of every variable $X$, in time $O(f)$, then we can solve for Corollaries 2.1 and 2.2 for all parameters in time $O(f)$. We do this by running the algorithm three times, once with $\mathbf{i} = \mathbf{e}$, and then with $\mathbf{i} = y, \mathbf{e}$, and finally with $\mathbf{i} = z, \mathbf{e}$.*

Recall that the family of a variable $X$ is the set containing $X$, and its parents $\mathbf{U}$ in the belief network.

The join-tree algorithm (Jensen, Lauritzen, & Olesen, 1990) and the differential approach (Darwiche, 2000) can both compute $Pr(\mathbf{i}, x, \mathbf{u})$, for a given instantiation $\mathbf{i}$ and all family instantiations $x, \mathbf{u}$ of every variable $X$ in $O(n \exp w)$ time. Here, $n$ is the number of variables in the belief network, and $w$ is the width of a given elimination order. Samiam uses the differential approach, and thus its running time to identify all possible parameter changes in a network is also $O(n \exp w)$. Note that this is also the time needed to answer one of the simplest queries, that of computing the probability of evidence $\mathbf{e}$.

## 2.2 Multi-Valued Variables

Our results can be easily extended to multi-valued variables, as long as we assume a model for changing co-varying parameters when one of them changes (Darwiche, 2000; Kjærulff & van der Gaag, 2000). After the parameter $\theta_{x|\mathbf{u}}$ changes, we need to use a scheme to change the other parameters, $\theta_{x_i|\mathbf{u}}$ for all $x_i \neq x$, in order to ensure the sum-to-one constraint. The most common way to do this is to use the proportional scheme. In this scheme, we change the other parameters so that the ratios between them remain the same. For example, suppose we have three parameters $\theta_{x_1|\mathbf{u}} = .6$, $\theta_{x_2|\mathbf{u}} = .3$ and $\theta_{x_3|\mathbf{u}} = .1$. After $\theta_{x_1|\mathbf{u}}$ changes to .8, the other two parameter values will be changed to $\theta_{x_2|\mathbf{u}} = .3(.2/.4) = .15$ and $\theta_{x_3|\mathbf{u}} = .1(.2/.4) = .05$ accordingly. We now define the meta parameter $\tau_{x|\mathbf{u}}$ such that it simultaneously changes all parameters according to the proportional scheme. We can then obtain a linear relation between $Pr(\mathbf{e})$ and $\tau_{x|\mathbf{u}}$, and the partial derivative is given by:

$$\frac{\partial Pr(\mathbf{e})}{\tau_{x|\mathbf{u}}} = \frac{Pr(\mathbf{e}, x, \mathbf{u})}{\theta_{x|\mathbf{u}}} - \frac{\sum_{x_i \neq x} Pr(\mathbf{e}, x_i, \mathbf{u})}{\sum_{x_i \neq x} \theta_{x_i|\mathbf{u}}}.$$

This is very similar to the result in Theorem 2.1, in the way that we have grouped all the values $x_i \neq x$ into the value $\overline{x}$. We can then use Corollaries 2.1 and 2.2 to solve for the Difference and Ratio constraints.

We now present another example to illustrate how the results above are used in practice.

**Example 2.1**  *Consider again the network in Figure 3. Here, we set the evidence such that we have smoke, but no report of people evacuating the building, i.e.* $\mathbf{e} = smoke, \overline{report}$*. We then got the posteriors* $Pr(fire \mid \mathbf{e}) = .25$ *and* $Pr(tampering \mid \mathbf{e}) = .02$*. We thought in this case that the posterior on fire should be no less than .5 and asked* SAMIAM *to recommend the necessary changes to enforce the constraint,* $Pr(fire \mid \mathbf{e}) - Pr(\overline{fire} \mid \mathbf{e}) \geq 0$*. There were five recommendations in this case, three of which could be ruled out based on qualitative considerations:*

1. *increase the prior on fire to* $\geq .03$ *(from .01);*

2. *increase the prior on tampering to* $\geq .80$ *(from .02);*

3. *decrease* $Pr(smoke \mid \overline{fire})$ *to* $\leq .003$ *(from .01);*

4. *increase* $Pr(leaving \mid \overline{alarm})$ *to* $\geq .923$ *(from .001);*

5. *increase* $Pr(report \mid \overline{leaving})$ *to* $\geq .776$ *(from .01).*

*Clearly, the only sensible change here is either to increase the prior on fire, or to decrease the probability of having smoke without a fire.*

This example and other similar ones suggest that identifying such parameter changes and their magnitudes is inevitable for developing a faithful belief network, yet it is not trivial for experts to accomplish this task by visual inspection of the belief network, often due to its size and complexity. Sensitivity analysis tools such as SAMIAM can help facilitate this by identifying important parameters that need to be fine-tuned in order to satisfy certain constraints. Of course, if we are given multiple constraints, we need to be cautious when implementing a recommendation made by SAMIAM due to one constraint, because this may result in violating other constraints. In this case, the parameter changes recommended by SAMIAM should be used to help experts in focusing their attention on the relevant parameters.

Moreover, the previous examples illustrate the need to develop more analytic tools to understand and explain the sensitivity of queries to certain parameter changes. There is also a need to reconcile the sensitivities exhibited by our examples with previous experimental studies demonstrating the robustness of probabilistic queries against small parameter changes in certain application areas, such as diagnosis (Pradhan et al., 1996). We address these particular questions in the next two sections.

## 3. The Sensitivity of Probabilistic Queries to Parameters Changes

Our starting point in understanding the sensitivity of a query $Pr(y \mid \mathbf{e})$ to changes in a meta parameter $\tau_{x|\mathbf{u}}$ is to analyze the derivative $\partial Pr(y \mid \mathbf{e})/\partial\tau_{x|\mathbf{u}}$. In our analysis, we assume that $X$ is binary, but $Y$ and all other variables in the network can be multi-valued. The following theorem provides a simple bound on this derivative, in terms of $Pr(y \mid \mathbf{e})$ and $Pr(x \mid \mathbf{u})$ only. We then use this simple bound to study the effect of changes to meta parameters on probabilistic queries.

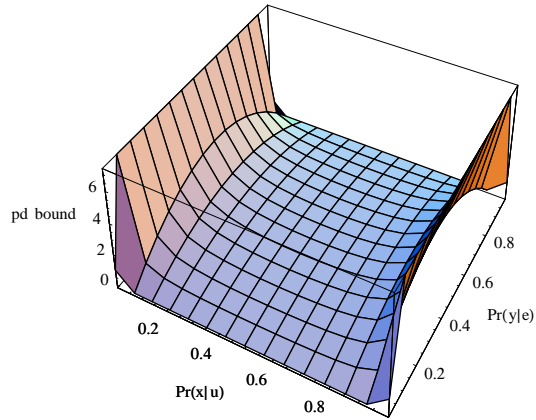

Figure 4: The plot of the upper bound on the partial derivative $\partial Pr(y \mid \mathbf{e})/\partial \tau_{x|\mathbf{u}}$, as given in Theorem 3.1, against $Pr(x \mid \mathbf{u})$ and $Pr(y \mid \mathbf{e})$.

**Theorem 3.1** *If $X$ is a binary variable in a belief network, then:*[4]

$$\left| \frac{\partial Pr(y \mid \mathbf{e})}{\partial \tau_{x|\mathbf{u}}} \right| \leq \frac{Pr(y \mid \mathbf{e})(1 - Pr(y \mid \mathbf{e}))}{Pr(x \mid \mathbf{u})(1 - Pr(x \mid \mathbf{u}))}.$$

The bound in Theorem 3.1 is tight, and we will show later an example for which the derivative assumes the above bound exactly. The main point to note about this bound is that it is independent of any given belief network.[5]

The plot of this bound against $Pr(x \mid \mathbf{u})$ and $Pr(y \mid \mathbf{e})$ is shown in Figure 4. A number of observations are in order about this plot:

- For extreme values of $Pr(x \mid \mathbf{u})$, the bound approaches infinity, and thus a small absolute change in the meta parameter $\tau_{x|\mathbf{u}}$ can have a big impact on the query $Pr(y \mid \mathbf{e})$.

- On the other hand, the bound approaches 0 for extreme values of the query $Pr(y \mid \mathbf{e})$. Therefore, a small absolute change in the meta parameter $\tau_{x|\mathbf{u}}$ will have a small effect on the absolute change in the query.

One of the implications of this result is that if we have a belief network where queries of interest $Pr(y \mid \mathbf{e})$ have extreme values, then such queries will be robust against small changes in network parameters. This of course assumes that robustness is understood to be a small change in the absolute value of the given query. Interestingly enough, if $y$ is a disease which is diagnosed by finding $\mathbf{e}$—that is, the probability $Pr(y \mid \mathbf{e})$ is quite high— then it is not surprising that such queries would be robust against small perturbations to network parameters. This seems to explain some of the results by Pradhan et al. (1996), where robustness have been confirmed for queries with $Pr(y \mid \mathbf{e}) \geq .9$.

---

4. This theorem and all results that follow requires that $\tau_{x|\mathbf{u}} \neq 0$ and $\tau_{x|\mathbf{u}} \neq 1$, since we can only use the expression in Equation 2.1 under these conditions.

5. Note that we have an exact closed form for the derivative $\partial Pr(y \mid \mathbf{e})/\partial \tau_{x|\mathbf{u}}$ (Darwiche, 2000; Greiner, Grove, & Schuurmans, 1997), but that form includes terms which are specific to the given belief network.

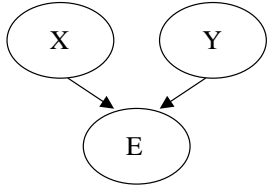

Figure 5: The network used in Example 3.1.

Another implication of the above result is that one has to be careful when changing parameters that are extreme. Such parameters are potentially very influential and one must handle them with care.

Therefore, the worst situation from a robustness viewpoint materializes if one has extreme parameters with non-extreme queries. In such a case, the queries can be very sensitive to small variations in the parameters.

**Example 3.1** *Consider the network structure in Figure 5. We have two binary nodes, $X$ and $Y$ with respective parameters $\theta_x, \theta_{\overline{x}}$ and $\theta_y, \theta_{\overline{y}}$. We assume that $E$ is a deterministic binary node where the value of $E$ is $e$ iff $X = Y$. This dictates the following CPT for $E$: $Pr(e \mid x, y) = 1$, $Pr(e \mid \overline{x}, \overline{y}) = 1$, $Pr(e \mid x, \overline{y}) = 0$ and $Pr(e \mid \overline{x}, y) = 0$. The conditional probability $Pr(y \mid e)$ can be expressed using the root parameters $\theta_x$ and $\theta_y$ as:*

$$Pr(y \mid e) = \frac{\theta_x \theta_y}{\theta_x \theta_y + \theta_{\overline{x}} \theta_{\overline{y}}}.$$

*Since $\partial\theta_x/\partial\tau_x = 1$ and $\partial\theta_{\overline{x}}/\partial\tau_x = -1$, the derivative of $Pr(y \mid e)$ with respect to the meta parameter $\tau_x$ is given by:*

$$
\begin{aligned}
\frac{\partial Pr(y \mid e)}{\partial \tau_x} &= \frac{(\theta_x \theta_y + \theta_{\overline{x}} \theta_{\overline{y}})\theta_y - \theta_x \theta_y(\theta_y - \theta_{\overline{y}})}{(\theta_x \theta_y + \theta_{\overline{x}} \theta_{\overline{y}})^2} \\
&= \frac{\theta_y \theta_{\overline{y}}}{(\theta_x \theta_y + \theta_{\overline{x}} \theta_{\overline{y}})^2}.
\end{aligned}
$$

*This is equal to the upper bound given in Theorem 3.1:*

$$
\begin{aligned}
\frac{Pr(y \mid e)(1 - Pr(y \mid e))}{Pr(x)(1 - Pr(x))} &= \frac{(\theta_x \theta_y)(\theta_{\overline{x}} \theta_{\overline{y}})}{\theta_x \theta_{\overline{x}}(\theta_x \theta_y + \theta_{\overline{x}} \theta_{\overline{y}})^2} \\
&= \frac{\theta_y \theta_{\overline{y}}}{(\theta_x \theta_y + \theta_{\overline{x}} \theta_{\overline{y}})^2}.
\end{aligned}
$$

*Now, if we set $\theta_x = \theta_{\overline{y}}$, the derivative becomes:*

$$\frac{\partial Pr(y \mid e)}{\partial \tau_x} = \frac{1}{4\theta_x \theta_{\overline{x}}},$$

and as $\theta_x$ (or $\theta_{\overline{x}}$) approaches 0, the derivative approaches infinity. Finally, if we set $\theta_x = \theta_{\overline{y}} = \epsilon$, we have $Pr(y \mid e) = .5$, but if we keep $\theta_y$ and $\theta_{\overline{y}}$ constant and change $\tau_x$ from $\epsilon$ to 0, we get the new result $Pr(y \mid e) = 0$.

Example 3.1 then illustrates three points. First, it shows that the bound in Theorem 3.1 is tight, i.e. we can construct a belief network that assumes the bound. Second, it gives an example network for which the derivative $\partial Pr(y \mid e)/\partial \tau_{x|\mathbf{u}}$ tends to infinity, and therefore we cannot bound the derivative by any constant. Finally, it shows that an infinitesimal absolute change in a meta parameter (changing $\tau_x$ from $\epsilon$ to 0) can induce a non-infinitesimal absolute change in some query ($Pr(y \mid e)$ changes from .5 to 0). The following theorem, however, shows that this is not possible if we consider a *relative* notion of change.

**Theorem 3.2** *Assume that $\tau_{x|\mathbf{u}} \leq .5$ without loss of generality.*[6] *Suppose that $\Delta \tau_{x|\mathbf{u}}$ is an infinitesimal change applied to the meta parameter $\tau_{x|\mathbf{u}}$, leading to a change of $\Delta Pr(y \mid \mathbf{e})$ to the query $Pr(y \mid \mathbf{e})$. We then have:*

$$\left| \frac{\Delta Pr(y \mid \mathbf{e})}{Pr(y \mid \mathbf{e})} \right| \leq 2 \left| \frac{\Delta \tau_{x|\mathbf{u}}}{\tau_{x|\mathbf{u}}} \right|.$$

For a function $f(x)$, the quantity:

$$\lim_{(x-x_0) \to 0} \frac{(f(x) - f(x_0))/f(x_0)}{(x - x_0)/x_0},$$

is typically known as the *sensitivity* of $f$ to $x$ at $x_0$. Therefore, Theorem 3.2 shows that the sensitivity of $Pr(y \mid \mathbf{e})$ to $\tau_{x|\mathbf{u}}$ is bounded.

As an example application of Theorem 3.2, consider Example 3.1 again. The change of $\tau_x$ from $\epsilon$ to 0 amounts to a relative change $|-\epsilon/\epsilon| = 1$. The corresponding change of $Pr(y \mid e)$ from .5 to 0 amounts to a relative change of $|-.5/.5| = 1$. Hence, the relative change in the query is not as great from this viewpoint.[7]

The relative change in $Pr(y \mid \mathbf{e})$ may be greater than double the relative change in $\tau_{x|\mathbf{u}}$ for non-infinitesimal changes because the derivative $\partial Pr(y \mid \mathbf{e})/\partial \tau_{x|\mathbf{u}}$ depends on the value of $\tau_{x|\mathbf{u}}$ (Darwiche, 2000; Jensen, 1999). Going back to Example 3.1, if we set $\theta_x = .5$ and $\theta_y = .01$, we obtain the result $Pr(y \mid e) = .01$. If we now increase $\tau_x$ to .6, a relative change of 20%, we get the new result $Pr(y \mid e) = 0.0149$, a relative change of 49%, which is more than double of the relative change in $\tau_x$.

The question now is: Suppose that we change a meta parameter $\tau_{x|\mathbf{u}}$ by an arbitrary amount (not an infinitesimal amount), what can we say about the corresponding change in the query $Pr(y \mid \mathbf{e})$? We have the following result.

**Theorem 3.3** *Let $O(x \mid \mathbf{u})$ denote the odds of $x$ given $\mathbf{u}$: $O(x \mid \mathbf{u}) = Pr(x \mid \mathbf{u})/(1 - Pr(x \mid \mathbf{u}))$, and let $O(y \mid \mathbf{e})$ denote the odds of $y$ given $\mathbf{e}$: $O(y \mid \mathbf{e}) = Pr(y \mid \mathbf{e})/(1 - Pr(y \mid \mathbf{e}))$. Let $O'(x \mid \mathbf{u})$ and $O'(y \mid \mathbf{e})$ denote these odds after having applied an arbitrary change to* the meta parameter $\tau_{x|\mathbf{u}}$ where $X$ is a binary variable in a belief network. If the change is positive, then:

$$\frac{O(x \mid \mathbf{u})}{O'(x \mid \mathbf{u})} \leq \frac{O'(y \mid \mathbf{e})}{O(y \mid \mathbf{e})} \leq \frac{O'(x \mid \mathbf{u})}{O(x \mid \mathbf{u})};$$

or if it is negative, then:

$$\frac{O'(x \mid \mathbf{u})}{O(x \mid \mathbf{u})} \leq \frac{O'(y \mid \mathbf{e})}{O(y \mid \mathbf{e})} \leq \frac{O(x \mid \mathbf{u})}{O'(x \mid \mathbf{u})}.$$

---

6. For a binary variable $X$, if $\tau_{x|\mathbf{u}} > .5$, we can instead choose the meta parameter $\tau_{\overline{x}|\mathbf{u}}$ without loss of generality.

7. If we consider the meta parameter $\tau_{\overline{x}} = 1 - \epsilon$ instead, the relative change in $\tau_{\overline{x}}$ will then amount to $\epsilon/(1 - \epsilon)$. But Theorem 3.2 will not be applicable in this case (assuming that $\epsilon$ is close to 0) since the theorem requires that the chosen meta parameter be no greater than .5.

Combining both results, we have:

$$|\ln(O'(y \mid \mathbf{e})) - \ln(O(y \mid \mathbf{e}))| \leq |\ln(O'(x \mid \mathbf{u})) - \ln(O(x \mid \mathbf{u}))|.$$

Theorem 3.3 means that the relative change in the odds of $y$ given $\mathbf{e}$ is bounded by the relative change in the odds of $x$ given $\mathbf{u}$, if $X$ is a binary variable.[8] *Note that the result makes no assumptions whatsoever about the structure of the given belief network.*

To illustrate this theorem, we go back to Example 2.1. We intend to increase the posterior $Pr(fire \mid \mathbf{e})$ from .25 to .5, for $\mathbf{e} = smoke, \overline{report}$. The log-odds change for the query is thus $\Delta lo(Pr(y \mid \mathbf{e})) = |\ln(O'(y \mid \mathbf{e})) - \ln(O(y \mid \mathbf{e}))| = 1.1$. There were five recommendations made by SamIam and we can calculate the log-odds change, $\Delta lo(\tau_{x|\mathbf{u}}) = |\ln(O'(x \mid \mathbf{u})) - \ln(O(x \mid \mathbf{u}))|$ for each parameter change:

1. increase the prior on fire to $\geq .03$ (from .01): $\Delta lo(\tau_{x|\mathbf{u}}) = 1.1$;

2. increase the prior on tampering to $\geq .80$ (from .02): $\Delta lo(\tau_{x|\mathbf{u}}) = 5.3$;

3. decrease $Pr(smoke \mid \overline{fire})$ to $\leq .003$ (from .01): $\Delta lo(\tau_{x|\mathbf{u}}) = 1.2$;

4. increase $Pr(leaving \mid \overline{alarm})$ to $\geq .923$ (from .001): $\Delta lo(\tau_{x|\mathbf{u}}) = 9.4$;

5. increase $Pr(report \mid \overline{leaving})$ to $\geq .776$ (from .01): $\Delta lo(\tau_{x|\mathbf{u}}) = 5.8$.

Therefore, we can see that all the recommended parameter changes satisfy Theorem 3.3, i.e. the log-odds change of the query is bounded by the log-odds change of the parameter.

An interesting special case of Theorem 3.3 is when $X$ is a root node and $X = Y$. From basic probability theory, we have:

$$O(x \mid \mathbf{e}) = O(x)\frac{Pr(\mathbf{e} \mid x)}{Pr(\mathbf{e} \mid \overline{x})}.$$

As the ratio $Pr(\mathbf{e} \mid x)/Pr(\mathbf{e} \mid \overline{x})$ is independent of $Pr(x)$, the ratio $O(x \mid \mathbf{e})/O(x)$ is also independent of this prior. Therefore, we can conclude that:

$$\frac{O'(x \mid \mathbf{e})}{O(x \mid \mathbf{e})} = \frac{O'(x)}{O(x)}. \tag{4}$$

This means we can find the exact amount of change needed for a meta parameter $\tau_x$ in order to induce a particular change on the query $Pr(x \mid \mathbf{e})$. There is no need to use the more expensive technique of Section 2 in this case.

---

8. We recently expanded our results to multi-valued variables, where we arbitrarily change parameters $\theta_{x|\mathbf{u}}$ to new values $\theta'_{x|\mathbf{u}}$, for all values $x$. The resulting bound is: $|\ln(O'(y \mid \mathbf{e})) - \ln(O(y \mid \mathbf{e}))| \leq \ln(\max_x \theta'_{x|\mathbf{u}}/\theta_{x|\mathbf{u}}) - \ln(\min_x \theta'_{x|\mathbf{u}}/\theta_{x|\mathbf{u}})$ (Chan & Darwiche, 2002).

**Example 3.2** *Consider the network in Figure 3. Suppose that* $\mathbf{e} = report, \overline{smoke}$. *Currently,* $Pr(tampering) = .02$ *and* $Pr(tampering \mid \mathbf{e}) = .50$. *We wish to increase the conditional probability to* .65. *We can compute the new prior probability* $Pr'(tampering)$ *using Equation 4:*

$$\frac{.65/.35}{.50/.50} = \frac{Pr'(tampering)/(1 - Pr'(tampering))}{.02/.98},$$

*giving us* $Pr'(tampering) = .036$, *which is equal to the result we obtained using* SamIam *in Section 1. Both the changes to* $Pr(tampering)$ *and* $Pr(tampering \mid \mathbf{e})$ *bring a log-odds difference of* .616.

Theorem 3.3 has a number of implications. First, given a particular query $Pr(y \mid \mathbf{e})$ and a meta parameter $\tau_{x|\mathbf{u}}$, it can be used to bound the effect that a change in $\tau_{x|\mathbf{u}}$ will have on the query $Pr(y \mid \mathbf{e})$. Going back to Example 3.2, we may wish to know what is the impact on other conditional probabilities if we apply the change making $Pr'(tampering) = .036$. The log-odds changes for all conditional probabilities in the network will be bounded by .616. For example, currently $Pr(fire \mid \mathbf{e}) = .029$. Using Theorem 3.3, we can find the range of the new conditional probability value $Pr'(fire \mid \mathbf{e})$:

$$\left| \ln\left( \frac{Pr'(fire \mid \mathbf{e})}{1 - Pr'(fire \mid \mathbf{e})} \right) - \ln\left( \frac{.029}{.971} \right) \right| \leq .616,$$

giving us the range $.016 \leq Pr'(fire \mid \mathbf{e}) \leq .053$. The exact value of $Pr'(fire \mid \mathbf{e})$, obtained by inference, is .021, which is within the computed bounds.

Second, Theorem 3.3 can be used to efficiently approximate solutions to the Difference and Ratio problems we discussed in Section 2. That is, given a desirable change in the value of query $Pr(y \mid \mathbf{e})$, we can use Theorem 3.3 to immediately compute a lower bound on the minimum change to meta parameter $\tau_{x|\mathbf{u}}$ needed to induce the change. This method can be applied in constant time and can serve as a preliminary recommendation, as the method proposed in Section 2 is much more expensive computationally.

Third, suppose that SamIam was used to recommend parameter changes that would induce a desirable change on a given query. Suppose further that SamIam returned a number of such changes, each of which is capable of inducing the necessary change. The question is: which one of these changes should we adopt? The main principle applied in these situations is to adopt a "minimal" change. But what is minimal in this case? As Theorem 3.3 reveals, a notion of minimality which is based on the amount of absolute change can be very misleading. Instead, it suggests that one adopts the change that minimizes the relative change in the odds, as other queries can be shown to be robust against such a change in a precise sense.

For example, we are given two parameter changes, one from .1 to .15, and another from .4 to .45. Both these changes give us the same absolute change of .05. However, the first change has an log-odds change of .462, while the second one has an log-odds change of .205. Therefore, two parameter changes that give us the same absolute change can have different amounts of log-odds change.

On the other hand, two parameter changes that give us the same relative change can also have different amounts of log-odds change. For example, we are given two parameter changes, one from .1 to .2, and another from .2 to .4. Both these changes double the original parameter value. However, the first change has a log-odds change of .811, while the second one has a log-odds change of .981.

Finally, the result can be used to obtain a better intuitive understanding of parameter changes that do or do not matter, a topic which we will discuss in the next section.

## 4. Changes that (Don't) Matter

We now return to a central question: When do changes in network parameters matter and when do they not matter? As we mentioned earlier, there have been experimental studies investigating the robustness of belief networks against parameter changes (Pradhan et al., 1996). But we have also shown very simple and intuitive examples where networks can be very sensitive to small parameter changes. This calls for a better understanding of the effect of parameter changes on queries, so one can intuitively sort out situations in which such changes do or do not matter. Our goal in this section is to further develop such an understanding by looking more closely into some of the implications of Theorem 3.3. We start first by highlighting the difference between this theorem and previous results on sensitivity analysis.

### 4.1 Network-Specific Sensitivity Analysis

One of the main differences between our results and other sensitivity analysis approaches is that we do not need to know the belief network, and hence, do not need to perform inference. To clarify this difference, we compare it with the sensitivity function approach (van der Gaag & Renooij, 2001), which computes the sensitivity function that relates a query, $f(x)$, and a parameter, $x$, in the form:

$$f(x) = \frac{a \cdot x + b}{c \cdot x + d},$$

where $a$, $b$, $c$, $d$ are constants that depend on the given network and are computed by performing inference as suggested by van der Gaag and Renooij (2001).

Going back to Example 2.1, we can express the query $Pr(fire \mid smoke, \overline{report})$ as a function of the parameter $x = Pr(smoke \mid \overline{fire})$. The function is given by:

$$f(x) = \frac{0.003165}{0.9684 \cdot x + 0.003165},$$

and we plot this function in Figure 6. We can see that at the current parameter value .01, the query value is .25, but if we decrease it to .003, the query value increases to .5, which is one of the suggested parameter changes by SAMIAM.

However, we can find a bound on the relations between the query and the parameter using Theorem 3.3, without doing inference on the network (and without knowing the network). For example, by changing the current parameter value from .01 to .003, the new query value will be within the bounds of .09 and .53. On the other hand, if we want the query value to increase to .5, we have to at least decrease the parameter value from .01 to .003, or increase it to .03.

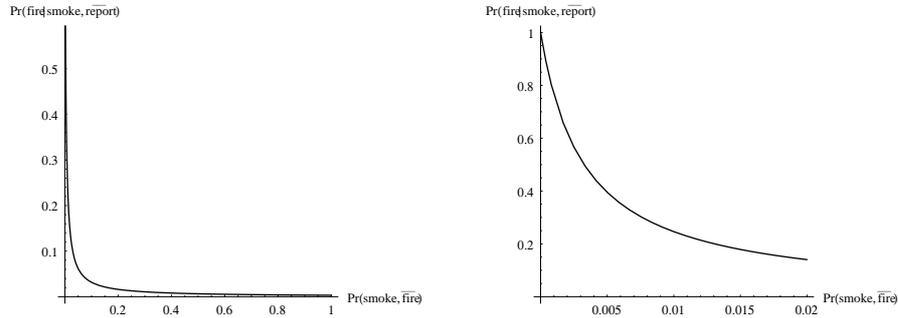

Figure 6: The plot of the query $Pr(\mathit{fire} \mid \mathit{smoke}, \overline{\mathit{report}})$ against the parameter $Pr(\mathit{smoke} \mid \overline{\mathit{fire}})$. The second graph shows a magnification of the first graph for the region where $Pr(\mathit{smoke} \mid \overline{\mathit{fire}})$ is between 0 and .02.

## 4.2 Assuring Query Robustness

One of the important issues we have yet to settle is: "What does it mean for a parameter change to *not* matter?" One can think of at least three definitions. First, the *absolute change* in the probability $Pr(y \mid \mathbf{e})$ is small. Second, the *relative change* in the probability $Pr(y \mid \mathbf{e})$ is small. Third, *relative change in the odds $O(y \mid \mathbf{e})$ is small*. The first notion is the one most prevalent in the literature, so we shall adopt it in the rest of this section.

Suppose we have a belief network for a diagnostic application and suppose we are concerned about the robustness of the query $Pr(y \mid \mathbf{e})$ with respect to changes in network parameters. In this application, $y$ is a particular disease and $\mathbf{e}$ is a particular finding which predicts the disease, with $Pr(y \mid \mathbf{e}) = .9$. Let us define robustness in this case to be an absolute change of no more than .05 to the given query. Now, let $X$ be a binary variable in the network and let us ask: What kind of changes to the parameters on $X$ are guaranteed to keep the query within the desirable range? We can use Theorem 3.3 easily to answer this question. First, if we are changing a parameter by $\delta$, and if we want the value of the query to remain $\leq .95$, we must ensure that:

$$|\ln((p + \delta)/(1 - p - \delta)) - \ln(p/(1 - p))| \leq |\ln(.95/.05) - \ln(.9/.1)| = .7472,$$

where $p$ is the current value of the parameter. Similarly, if we want to ensure that the query remains $\geq .85$, we want to ensure that:

$$|\ln((p + \delta)/(1 - p - \delta)) - \ln(p/(1 - p))| \leq |\ln(.85/.15) - \ln(.9/.1)| = .4626.$$

Figure 7 plots the permissible change $\delta$ as a function of $p$, the current value of the parameter. The main point to observe here is that the amount of permissible change depends on the current value of $p$, with smaller changes allowed for extreme values of $p$. It is also interesting to note that it is easier to guarantee the query to stay $\leq .95$ than to guarantee that it stays $\geq .85$. In general, it is more likely for a parameter change to reduce the value of a query which is close to 1 (and to increase the value of a query which is close to 0). Finally, if we are increasing the parameter, then a parameter value close to .4 will allow the biggest absolute change. But if we are decreasing the parameter, then a value close to .6 will allow the biggest absolute change.

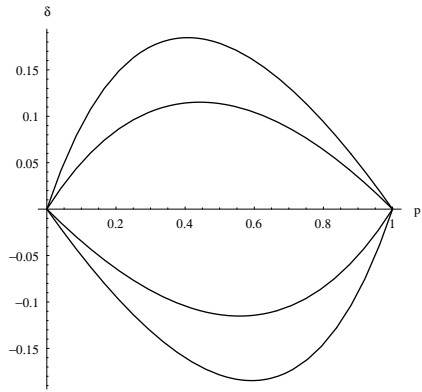

Figure 7: The amount of parameter change $\delta$ that would guarantee the query $Pr(y \mid \mathbf{e}) = .9$ to stay within the interval $[.85, .95]$, as a function of the current parameter value $p$. The outer envelope guarantees the query to remain $\leq .95$, while the inner envelope guarantees the query to remain $\geq .85$.

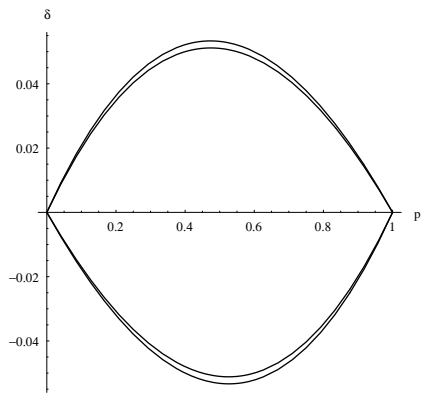

Figure 8: The amount of parameter change $\delta$ that would guarantee the query $Pr(y \mid \mathbf{e}) = .6$ to stay within the interval $[.55, .65]$, as a function of the current parameter value $p$. The outer envelope guarantees the query to remain $\leq .65$, while the inner envelope guarantees the query to stay in $\geq .55$.

Now let us repeat the same exercise but assuming that the initial value of the query is $Pr(y \mid \mathbf{e}) = .6$, yet insisting on the same measure of robustness. Figure 8 plots the permissible changes $\delta$ as a function of $p$, the current value of the parameter. Again, the amount of permissible change becomes smaller as the probability $p$ approaches 0 or 1. The other main point to emphasize is that the permissible changes are now much smaller than in the previous example, since the initial value of the query is not as extreme. Therefore, this query is much less robust than the previous one.

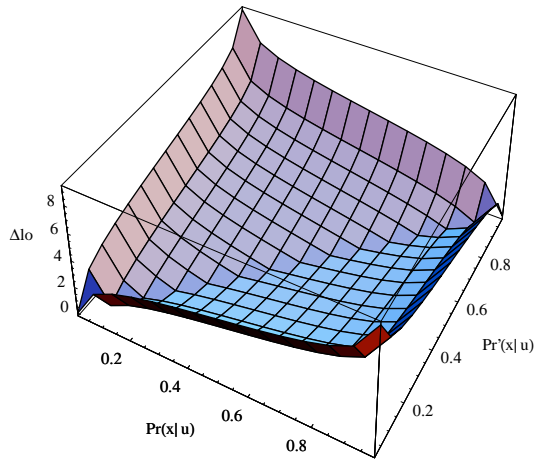

Figure 9: The plot of the log-odd difference, $\Delta lo = |\ln(O'(x \mid \mathbf{u})) - \ln(O(x \mid \mathbf{u}))|$, against $Pr(x \mid \mathbf{u})$ and $Pr'(x \mid \mathbf{u})$.

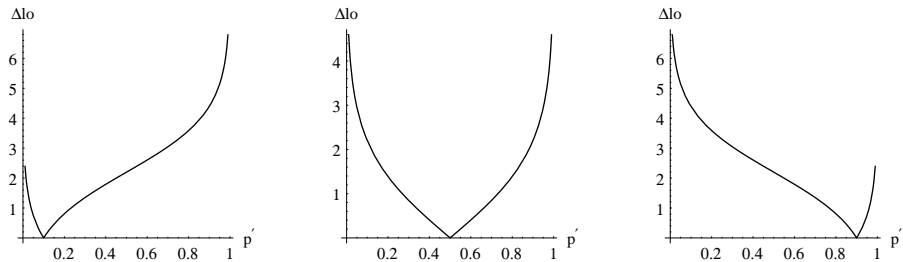

Figure 10: The plots of the log-odd difference, $\Delta lo = |\ln(O'(x \mid \mathbf{u})) - \ln(O(x \mid \mathbf{u}))|$, against the new parameter value $p' = Pr'(x \mid \mathbf{u})$. The figures correspond to different initial values of the parameter, $p = Pr(x \mid \mathbf{u}) = .1, .5, .9$, respectively.

More generally, Figure 9 plots the log-odds difference, $|\ln(O'(x \mid \mathbf{u})) - \ln(O(x \mid \mathbf{u}))|$, against $Pr(x \mid \mathbf{u}) = p$ and $Pr'(x \mid \mathbf{u}) = p + \delta$, and Figure 10 shows cross-sections of Figure 9 for three different values of $p$. Again, the plots explain analytically why we can afford more absolute changes to non-extreme probabilities (Pradhan et al., 1996; Poole, 1998).

From Figure 10, we also notice that although the plot is symmetric for $p = .5$, it is not for both $p = .1$ and $p = .9$, i.e. absolute changes of $\Delta p$ and $-\Delta p$ give us different amounts of log-odds change. For example, changing the parameter from .1 to .05 give us a larger log-odds change than changing the parameter from .1 to .15. We also notice that the plots for $p = .1$ and $p = .9$ are mirror images of each other. Therefore, the log-odds change is the same for complementary parameter changes on $\theta_{x|\mathbf{u}}$ and $\theta_{\overline{x}|\mathbf{u}}$.

We close this section by emphasizing that the above figures identify parameter changes that guarantee keeping queries within certain ranges. However, if the belief network has specific properties, such as a specific topology, then it is possible for the query to be robust against parameter changes that are outside the identified bounds.

## 5. Conclusion

In this paper, we presented an efficient technique for fine-tuning the parameters of a belief network. The technique suggests minimal changes to network parameters which ensure that certain constraints are enforced on probabilistic queries. Based on this technique, we have experimented with some belief networks, only to find out that these networks are more sensitive to parameter changes than previous experimental studies seem to suggest. This observation leads us to an analytic study on the effect of parameter changes, with the aim of characterizing situations under which parameter changes do or do not matter. We have reported on a number of results in this direction. Our central result shows that belief networks are robust in a very specific sense: the relative change in query odds is bounded by the relative change in the parameter odds. A closer look at this result, its meaning, and its implications provides interesting characterizations of parameter changes that do or do not matter, and explains analytically some of the previous experimental results and observations on this matter.

## Acknowledgments

A shorter version of this paper appeared in *Proceedings of the 17th Conference on Uncertainty in Artificial Intelligence (UAI-01)*, pp. 65–74. This work has been partially supported by NSF grant IIS-9988543, MURI grant N00014-00-1-0617, and by DiMI grant 00-10065.

## Appendix A. Proofs

**Theorem 2.1** *The derivative of $Pr(\mathbf{e})$ with respect to the meta parameter $\tau_{x|\mathbf{u}}$ is given by:*

$$\frac{\partial Pr(\mathbf{e})}{\partial \tau_{x|\mathbf{u}}} = \frac{Pr(\mathbf{e}, x, \mathbf{u})}{\theta_{x|\mathbf{u}}} - \frac{Pr(\mathbf{e}, \overline{x}, \mathbf{u})}{\theta_{\overline{x}|\mathbf{u}}},$$

*when $\theta_{x|\mathbf{u}} \neq 0$ and $\theta_{\overline{x}|\mathbf{u}} \neq 0$.*

**Proof** From Russell et al. (1995), the semantics of the first derivative of $Pr(\mathbf{e})$ with respect to parameter $\theta_{x|\mathbf{u}}$ is given by:[9]

$$\frac{\partial Pr(\mathbf{e})}{\partial \theta_{x|\mathbf{u}}} = \frac{Pr(\mathbf{e}, x, \mathbf{u})}{\theta_{x|\mathbf{u}}},$$

if $\theta_{x|\mathbf{u}} \neq 0$, and:

$$\frac{\partial Pr(\mathbf{e})}{\partial \theta_{\overline{x}|\mathbf{u}}} = \frac{Pr(\mathbf{e}, \overline{x}, \mathbf{u})}{\theta_{\overline{x}|\mathbf{u}}},$$

if $\theta_{\overline{x}|\mathbf{u}} \neq 0$. Because $\theta_{x|\mathbf{u}} = \tau_{x|\mathbf{u}}$ and $\theta_{\overline{x}|\mathbf{u}} = 1 - \tau_{x|\mathbf{u}}$, we have:

$$\begin{aligned}
\frac{\partial Pr(\mathbf{e})}{\partial \tau_{x|\mathbf{u}}} &= \frac{\partial Pr(\mathbf{e})}{\partial \theta_{x|\mathbf{u}}} - \frac{\partial Pr(\mathbf{e})}{\partial \theta_{\overline{x}|\mathbf{u}}} \\
&= \frac{Pr(\mathbf{e}, x, \mathbf{u})}{\theta_{x|\mathbf{u}}} - \frac{Pr(\mathbf{e}, \overline{x}, \mathbf{u})}{\theta_{\overline{x}|\mathbf{u}}},
\end{aligned}$$

if $\theta_{x|\mathbf{u}} \neq 0$ and $\theta_{\overline{x}|\mathbf{u}} \neq 0$. $\square$

**Theorem 3.1** *If $X$ is a binary variable in a belief network, then:*

$$\left| \frac{\partial Pr(y \mid \mathbf{e})}{\partial \tau_{x|\mathbf{u}}} \right| \leq \frac{Pr(y \mid \mathbf{e})(1 - Pr(y \mid \mathbf{e}))}{Pr(x \mid \mathbf{u})(1 - Pr(x \mid \mathbf{u}))}.$$

**Proof** From Darwiche (2000), the derivative $\partial Pr(y \mid \mathbf{e}) / \partial \theta_{x|\mathbf{u}}$ is equal to:

$$\frac{\partial Pr(y \mid \mathbf{e})}{\partial \theta_{x|\mathbf{u}}} = \frac{Pr(y, x, \mathbf{u} \mid \mathbf{e}) - Pr(y \mid \mathbf{e})Pr(x, \mathbf{u} \mid \mathbf{e})}{Pr(x \mid \mathbf{u})}.$$

Since:

$$\frac{\partial Pr(y \mid \mathbf{e})}{\partial \tau_{x|\mathbf{u}}} = \frac{\partial Pr(y \mid \mathbf{e})}{\partial \theta_{x|\mathbf{u}}} - \frac{\partial Pr(y \mid \mathbf{e})}{\partial \theta_{\overline{x}|\mathbf{u}}},$$

we have:

$$\begin{aligned}
&\frac{\partial Pr(y \mid \mathbf{e})}{\partial \tau_{x|\mathbf{u}}} \\
&= \frac{Pr(y, x, \mathbf{u} \mid \mathbf{e}) - Pr(y \mid \mathbf{e})Pr(x, \mathbf{u} \mid \mathbf{e})}{Pr(x \mid \mathbf{u})} - \frac{Pr(y, \overline{x}, \mathbf{u} \mid \mathbf{e}) - Pr(y \mid \mathbf{e})Pr(\overline{x}, \mathbf{u} \mid \mathbf{e})}{Pr(\overline{x} \mid \mathbf{u})} \\
&= \frac{Pr(y, x, \mathbf{u} \mid \mathbf{e}) - Pr(y \mid \mathbf{e})Pr(x, \mathbf{u} \mid \mathbf{e}) - Pr(x \mid \mathbf{u})(Pr(y, \mathbf{u} \mid \mathbf{e}) - Pr(y \mid \mathbf{e})Pr(\mathbf{u} \mid \mathbf{e}))}{Pr(x \mid \mathbf{u})(1 - Pr(x \mid \mathbf{u}))}.
\end{aligned}$$

In order to find an upper bound on the derivative, we would like to bound the term $Pr(y, x, \mathbf{u} \mid \mathbf{e}) - Pr(y \mid \mathbf{e})Pr(x, \mathbf{u} \mid \mathbf{e})$. Since, $Pr(y, x, \mathbf{u}, \mathbf{e}) \leq Pr(y, \mathbf{u}, \mathbf{e})$ and $Pr(y, x, \mathbf{u}, \mathbf{e}) \leq Pr(x, \mathbf{u}, \mathbf{e})$, we have:

$$\begin{aligned}
Pr(y, x, \mathbf{u} \mid \mathbf{e}) - Pr(y \mid \mathbf{e})Pr(x, \mathbf{u} \mid \mathbf{e}) &\leq Pr(y, x, \mathbf{u} \mid \mathbf{e}) - Pr(y \mid \mathbf{e})Pr(y, x, \mathbf{u} \mid \mathbf{e}) \\
&= Pr(y, x, \mathbf{u} \mid \mathbf{e})Pr(\overline{y} \mid \mathbf{e}) \\
&\leq Pr(y, \mathbf{u} \mid \mathbf{e})Pr(\overline{y} \mid \mathbf{e}).
\end{aligned}$$

---

9. We allow the notations $\partial Pr(\mathbf{e}) / \partial \theta_{x|\mathbf{u}}$ and $\partial Pr(\mathbf{e}) / \partial \theta_{\overline{x}|\mathbf{u}}$ by assuming $Pr(\mathbf{e})$ as functions of $\theta_{x|\mathbf{u}}$ and $\theta_{\overline{x}|\mathbf{u}}$, even though it is not allowed in belief networks to change only $\theta_{x|\mathbf{u}}$ or $\theta_{\overline{x}|\mathbf{u}}$.

Therefore, the upper bound on the derivative is given by:

$$\frac{\partial Pr(y \mid \mathbf{e})}{\partial \tau_{x|\mathbf{u}}} \le \frac{Pr(y, \mathbf{u} \mid \mathbf{e})Pr(\overline{y} \mid \mathbf{e}) - Pr(x \mid \mathbf{u})(Pr(y, \mathbf{u} \mid \mathbf{e}) - Pr(y \mid \mathbf{e})Pr(\mathbf{u} \mid \mathbf{e}))}{Pr(x \mid \mathbf{u})(1 - Pr(x \mid \mathbf{u}))},$$

which is equal to the following term:

$$
\begin{aligned}
& \frac{Pr(\overline{y} \mid \mathbf{e})Pr(y, \mathbf{u} \mid \mathbf{e})}{Pr(x \mid \mathbf{u})} + \frac{Pr(y \mid \mathbf{e})Pr(\overline{y}, \mathbf{u} \mid \mathbf{e})}{1 - Pr(x \mid \mathbf{u})} \\
= & \frac{(1 - Pr(x \mid \mathbf{u}))Pr(\overline{y} \mid \mathbf{e})Pr(y, \mathbf{u} \mid \mathbf{e}) + Pr(x \mid \mathbf{u})Pr(y \mid \mathbf{e})Pr(\overline{y}, \mathbf{u} \mid \mathbf{e})}{Pr(x \mid \mathbf{u})(1 - Pr(x \mid \mathbf{u}))} \\
= & \frac{Pr(y, \mathbf{u} \mid \mathbf{e})Pr(\overline{y} \mid \mathbf{e}) - Pr(x \mid \mathbf{u})(Pr(y, \mathbf{u} \mid \mathbf{e}) - Pr(y \mid \mathbf{e})Pr(\mathbf{u} \mid \mathbf{e}))}{Pr(x \mid \mathbf{u})(1 - Pr(x \mid \mathbf{u}))}.
\end{aligned}
$$

Since $Pr(y, \mathbf{u} \mid \mathbf{e}) \le Pr(y \mid \mathbf{e})$ and $Pr(\overline{y}, \mathbf{u} \mid \mathbf{e}) \le Pr(\overline{y} \mid \mathbf{e})$, the upper bound on the derivative is given by:

$$
\begin{aligned}
\frac{\partial Pr(y \mid \mathbf{e})}{\partial \tau_{x|\mathbf{u}}} & \le \frac{Pr(\overline{y} \mid \mathbf{e})Pr(y, \mathbf{u} \mid \mathbf{e})}{Pr(x \mid \mathbf{u})} + \frac{Pr(y \mid \mathbf{e})Pr(\overline{y}, \mathbf{u} \mid \mathbf{e})}{1 - Pr(x \mid \mathbf{u})} \\
& \le \frac{Pr(\overline{y} \mid \mathbf{e})Pr(y \mid \mathbf{e})}{Pr(x \mid \mathbf{u})} + \frac{Pr(y \mid \mathbf{e})Pr(\overline{y} \mid \mathbf{e})}{1 - Pr(x \mid \mathbf{u})} \\
& = \frac{Pr(y \mid \mathbf{e})(1 - Pr(y \mid \mathbf{e}))}{Pr(x \mid \mathbf{u})(1 - Pr(x \mid \mathbf{u}))}.
\end{aligned}
$$

In order to find a lower bound on the derivative, we note that $Pr(\overline{y} \mid \mathbf{e}) = 1 - Pr(y \mid \mathbf{e})$, and thus $\partial Pr(\overline{y} \mid \mathbf{e})/\partial \tau_{x|\mathbf{u}} = -\partial Pr(y \mid \mathbf{e})/\partial \tau_{x|\mathbf{u}}$. Therefore, we can get our lower bound by finding the upper bound on the derivative $\partial Pr(\overline{y} \mid \mathbf{e})/\partial \tau_{x|\mathbf{u}}$ and multiplying by $-1$:

$$
\begin{aligned}
\frac{\partial Pr(y \mid \mathbf{e})}{\partial \tau_{x|\mathbf{u}}} & \ge -\frac{Pr(\overline{y} \mid \mathbf{e})(1 - Pr(\overline{y} \mid \mathbf{e}))}{Pr(x \mid \mathbf{u})(1 - Pr(x \mid \mathbf{u}))} \\
& = -\frac{Pr(y \mid \mathbf{e})(1 - Pr(y \mid \mathbf{e}))}{Pr(x \mid \mathbf{u})(1 - Pr(x \mid \mathbf{u}))}.
\end{aligned}
$$

Combining the upper bound and the lower bound, we have:

$$\left| \frac{\partial Pr(y \mid \mathbf{e})}{\partial \tau_{x|\mathbf{u}}} \right| \le \frac{Pr(y \mid \mathbf{e})(1 - Pr(y \mid \mathbf{e}))}{Pr(x \mid \mathbf{u})(1 - Pr(x \mid \mathbf{u}))}. \quad \square$$

**Theorem 3.2** *Assume that $\tau_{x|\mathbf{u}} \le .5$ without loss of generality. Suppose that $\Delta \tau_{x|\mathbf{u}}$ is an infinitesimal change applied to the meta parameter $\tau_{x|\mathbf{u}}$, leading to a change of $\Delta Pr(y \mid \mathbf{e})$ to the query $Pr(y \mid \mathbf{e})$. We then have:*

$$\left| \frac{\Delta Pr(y \mid \mathbf{e})}{Pr(y \mid \mathbf{e})} \right| \le 2 \left| \frac{\Delta \tau_{x|\mathbf{u}}}{\tau_{x|\mathbf{u}}} \right|.$$

**Proof**  Because $\Delta\tau_{x|\mathbf{u}}$ is infinitesimal, from Theorem 3.1:

$$\left|\frac{\Delta Pr(y\mid\mathbf{e})}{\Delta\tau_{x|\mathbf{u}}}\right| \simeq \left|\frac{\partial Pr(y\mid\mathbf{e})}{\partial\tau_{x|\mathbf{u}}}\right|$$

$$\leq \frac{Pr(y\mid\mathbf{e})(1-Pr(y\mid\mathbf{e}))}{Pr(x\mid\mathbf{u})(1-Pr(x\mid\mathbf{u}))}.$$

Arranging the terms, we have:

$$\left|\frac{\Delta Pr(y\mid\mathbf{e})}{Pr(y\mid\mathbf{e})}\right| \leq \frac{1-Pr(y\mid\mathbf{e})}{1-Pr(x\mid\mathbf{u})}\left|\frac{\Delta\tau_{x|\mathbf{u}}}{\tau_{x|\mathbf{u}}}\right|$$

$$\leq \frac{1}{.5}\left|\frac{\Delta\tau_{x|\mathbf{u}}}{\tau_{x|\mathbf{u}}}\right|$$

$$= 2\left|\frac{\Delta\tau_{x|\mathbf{u}}}{\tau_{x|\mathbf{u}}}\right|,$$

since $Pr(x\mid\mathbf{u})=\tau_{x|\mathbf{u}}\leq .5$. $\square$

**Theorem 3.3**  *Let $O(x\mid\mathbf{u})$ denote the odds of $x$ given $\mathbf{u}$: $O(x\mid\mathbf{u})=Pr(x\mid\mathbf{u})/(1-Pr(x\mid\mathbf{u}))$, and let $O(y\mid\mathbf{e})$ denote the odds of $y$ given $\mathbf{e}$: $O(y\mid\mathbf{e})=Pr(y\mid\mathbf{e})/(1-Pr(y\mid\mathbf{e}))$. Let $O'(x\mid\mathbf{u})$ and $O'(y\mid\mathbf{e})$ denote these odds after having applied an arbitrary change to the meta parameter $\tau_{x|\mathbf{u}}$ where $X$ is a binary variable in a belief network. If the change is positive, then:*

$$\frac{O(x\mid\mathbf{u})}{O'(x\mid\mathbf{u})}\leq\frac{O'(y\mid\mathbf{e})}{O(y\mid\mathbf{e})}\leq\frac{O'(x\mid\mathbf{u})}{O(x\mid\mathbf{u})};$$

*or if it is negative, then:*

$$\frac{O'(x\mid\mathbf{u})}{O(x\mid\mathbf{u})}\leq\frac{O'(y\mid\mathbf{e})}{O(y\mid\mathbf{e})}\leq\frac{O(x\mid\mathbf{u})}{O'(x\mid\mathbf{u})}.$$

*Combining both results, we have:*

$$|\ln(O'(y\mid\mathbf{e}))-\ln(O(y\mid\mathbf{e}))|\leq |\ln(O'(x\mid\mathbf{u}))-\ln(O(x\mid\mathbf{u}))|.$$

**Proof**  We obtain this result by integrating the bound in Theorem 3.1. In particular, if we change $\tau_{x|\mathbf{u}}$ to $\tau'_{x|\mathbf{u}}>\tau_{x|\mathbf{u}}$, and consequently $Pr(y\mid\mathbf{e})$ changes to $Pr'(y\mid\mathbf{e})$, we can separate the variables in the upper bound on the derivative in Theorem 3.1, integrate over the intervals, and yield:

$$\int_{Pr(y|\mathbf{e})}^{Pr'(y|\mathbf{e})}\frac{dPr(y\mid\mathbf{e})}{Pr(y\mid\mathbf{e})(1-Pr(y\mid\mathbf{e}))}\leq\int_{\tau_{x|\mathbf{u}}}^{\tau'_{x|\mathbf{u}}}\frac{d\tau_{x|\mathbf{u}}}{\tau_{x|\mathbf{u}}(1-\tau_{x|\mathbf{u}})}.$$

This gives us the solution:

$$\ln(Pr'(y\mid\mathbf{e}))-\ln(Pr(y\mid\mathbf{e}))-\ln(1-Pr'(y\mid\mathbf{e}))+\ln(1-Pr(y\mid\mathbf{e}))$$
$$\leq \ln(\tau'_{x|\mathbf{u}})-\ln(\tau_{x|\mathbf{u}})-\ln(1-\tau'_{x|\mathbf{u}})+\ln(1-\tau_{x|\mathbf{u}}),$$

and after taking exponentials, we have:

$$\frac{Pr'(y \mid \mathbf{e})/(1 - Pr'(y \mid \mathbf{e}))}{Pr(y \mid \mathbf{e})/(1 - Pr(y \mid \mathbf{e}))} \leq \frac{\tau'_{x|\mathbf{u}}/(1 - \tau'_{x|\mathbf{u}})}{\tau_{x|\mathbf{u}}/(1 - \tau_{x|\mathbf{u}})},$$

which is equivalent to:

$$\frac{O'(y \mid \mathbf{e})}{O(y \mid \mathbf{e})} \leq \frac{O'(x \mid \mathbf{u})}{O(x \mid \mathbf{u})}.$$

Similarly, we can separate the variables in the lower bound on the derivative in Theorem 3.1, integrate over the intervals, and yield:

$$\int_{Pr(y|\mathbf{e})}^{Pr'(y|\mathbf{e})} \frac{dPr(y \mid \mathbf{e})}{Pr(y \mid \mathbf{e})(1 - Pr(y \mid \mathbf{e}))} \geq -\int_{\tau_{x|\mathbf{u}}}^{\tau'_{x|\mathbf{u}}} \frac{d\tau_{x|\mathbf{u}}}{\tau_{x|\mathbf{u}}(1 - \tau_{x|\mathbf{u}})}.$$

This gives us the solution:

$$\ln(Pr'(y \mid \mathbf{e})) - \ln(Pr(y \mid \mathbf{e})) - \ln(1 - Pr'(y \mid \mathbf{e})) + \ln(1 - Pr(y \mid \mathbf{e}))$$
$$\geq -\ln(\tau'_{x|\mathbf{u}}) + \ln(\tau_{x|\mathbf{u}}) + \ln(1 - \tau'_{x|\mathbf{u}}) - \ln(1 - \tau_{x|\mathbf{u}}),$$

and after taking exponentials, we have:

$$\frac{Pr'(y \mid \mathbf{e})/(1 - Pr'(y \mid \mathbf{e}))}{Pr(y \mid \mathbf{e})/(1 - Pr(y \mid \mathbf{e}))} \geq \frac{\tau_{x|\mathbf{u}}/(1 - \tau_{x|\mathbf{u}})}{\tau'_{x|\mathbf{u}}/(1 - \tau'_{x|\mathbf{u}})},$$

which is equivalent to:

$$\frac{O'(y \mid \mathbf{e})}{O(y \mid \mathbf{e})} \geq \frac{O(x \mid \mathbf{u})}{O'(x \mid \mathbf{u})}.$$

Therefore, we have the following inequality if $\tau'_{x|\mathbf{u}} > \tau_{x|\mathbf{u}}$:

$$\frac{O(x \mid \mathbf{u})}{O'(x \mid \mathbf{u})} \leq \frac{O'(y \mid \mathbf{e})}{O(y \mid \mathbf{e})} \leq \frac{O'(x \mid \mathbf{u})}{O(x \mid \mathbf{u})}.$$

On the other hand, if we now change $\tau_{x|\mathbf{u}}$ to $\tau'_{x|\mathbf{u}} < \tau_{x|\mathbf{u}}$, we can instead integrate from $\tau'_{x|\mathbf{u}}$ to $\tau_{x|\mathbf{u}}$. The integrals will satisfy these two inequalities:

$$\int_{Pr'(y|\mathbf{e})}^{Pr(y|\mathbf{e})} \frac{dPr(y \mid \mathbf{e})}{Pr(y \mid \mathbf{e})(1 - Pr(y \mid \mathbf{e}))} \leq \int_{\tau'_{x|\mathbf{u}}}^{\tau_{x|\mathbf{u}}} \frac{d\tau_{x|\mathbf{u}}}{\tau_{x|\mathbf{u}}(1 - \tau_{x|\mathbf{u}})};$$
$$\int_{Pr'(y|\mathbf{e})}^{Pr(y|\mathbf{e})} \frac{dPr(y \mid \mathbf{e})}{Pr(y \mid \mathbf{e})(1 - Pr(y \mid \mathbf{e}))} \geq -\int_{\tau'_{x|\mathbf{u}}}^{\tau_{x|\mathbf{u}}} \frac{d\tau_{x|\mathbf{u}}}{\tau_{x|\mathbf{u}}(1 - \tau_{x|\mathbf{u}})}.$$

We can solve for the inequalities similarly and get the result:

$$\frac{O'(x \mid \mathbf{u})}{O(x \mid \mathbf{u})} \leq \frac{O'(y \mid \mathbf{e})}{O(y \mid \mathbf{e})} \leq \frac{O(x \mid \mathbf{u})}{O'(x \mid \mathbf{u})}.$$

Combining the results for both $\tau'_{x|\mathbf{u}} > \tau_{x|\mathbf{u}}$ and $\tau'_{x|\mathbf{u}} < \tau_{x|\mathbf{u}}$, we have:

$$\mid \ln(O'(y \mid \mathbf{e})) - \ln(O(y \mid \mathbf{e})) \mid \leq \ \mid \ln(O'(x \mid \mathbf{u})) - \ln(O(x \mid \mathbf{u})) \mid. \square$$

# References

Castillo, E., Gutiérrez, J. M., & Hadi, A. S. (1997). Sensitivity analysis in discrete Bayesian networks. *IEEE Transactions on Systems, Man, and Cybernetics*, *27*, 412–423.

Chan, H., & Darwiche, A. (2002). A distance measure for bounding probabilistic belief change. In *Proceedings of the Eighteenth National Conference on Artificial Intelligence (AAAI)*, pp. 539–545.

Coupé, V. M. H., Peek, N., Ottenkamp, J., & Habbema, J. D. F. (1999). Using sensitivity analysis for efficient quantification of a belief network. *Artificial Intelligence in Medicine*, *17*, 223–247.

Darwiche, A. (2000). A differential approach to inference in Bayesian networks. In *Proceedings of the 16th Conference on Uncertainty in Artificial Intelligence (UAI)*, pp. 123–132.

Greiner, R., Grove, A., & Schuurmans, D. (1997). Learning Bayesian nets that perform well. In *Proceedings of the 13th Conference on Uncertainty in Artificial Intelligence (UAI)*, pp. 198–207.

Jensen, F. V., Lauritzen, S., & Olesen, K. (1990). Bayesian updating in recursive graphical models by local computation. *Computational Statistics Quarterly*, *4*, 269–282.

Jensen, F. V. (1999). Gradient descent training of bayesian networks. In *Proceedings of the Fifth European Conference on Symbolic and Quantitative Approaches to Reasoning with Uncertainty (ECSQARU)*, pp. 190–200.

Jensen, F. V. (2001). *Bayesian Networks and Decision Graphs*. Springer-Verlag, Inc., New York.

Kjærulff, U., & van der Gaag, L. C. (2000). Making sensitivity analysis computationally efficient. In *Proceedings of the 16th Conference on Uncertainty in Artificial Intelligence (UAI)*, pp. 317–325.

Laskey, K. B. (1995). Sensitivity analysis for probability assessments in Bayesian networks. *IEEE Transactions on Systems, Man, and Cybernetics*, *25*, 901–909.

Pearl, J. (1988). *Probabilistic Reasoning in Intelligent Systems: Networks of Plausible Inference*. Morgan Kaufmann Publishers, Inc., San Mateo, California.

Poole, D. (1998). Context-specific approximation in probabilistic inference. In *Proceedings of the 14th Conference on Uncertainty in Artificial Intelligence (UAI)*, pp. 447–454.

Pradhan, M., Henrion, M., Provan, G., Del Favero, B., & Huang, K. (1996). The sensitivity of belief networks to imprecise probabilities: an experimental investigation. *Artificial Intelligence*, *85*, 363–397.

Russell, S., Binder, J., Koller, D., & Kanazawa, K. (1995). Local learning in probabilistic networks with hidden variables. In *Proceedings of the Fourteenth International Joint Conference on Artificial Intelligence (IJCAI)*, pp. 1146–1152.

van der Gaag, L. C., & Renooij, S. (2001). Analysing sensitivity data from probabilistic networks. In *Proceedings of the 17th Conference on Uncertainty in Artificial Intelligence (UAI)*, pp. 530–537.